\documentclass{article}

\PassOptionsToPackage{numbers}{natbib}

\usepackage[eandd, final]{neurips_2026}

\usepackage{hyperref}
\usepackage{url}
\usepackage{enumitem}
\usepackage{graphicx}
\usepackage{setspace}
\usepackage{algorithmic}
\usepackage{algorithm}
\usepackage{amsmath}
\usepackage{amssymb}
\usepackage{mathtools}
\usepackage{amsthm}
\usepackage{fix-cm}
\usepackage{booktabs}
\usepackage{adjustbox}
\usepackage{colortbl}
\usepackage{xcolor}
\usepackage{diagbox}
\usepackage{makecell}
\usepackage{caption}
\usepackage{multirow}
\usepackage{multicol}
\usepackage{wrapfig}
\usepackage{graphicx}
\usepackage{subcaption}
\usepackage{nccmath}

\usepackage[capitalize,noabbrev]{cleveref}

\theoremstyle{plain}

\theoremstyle{definition}

\theoremstyle{remark}

\usepackage[textsize=tiny]{todonotes}

\setlist[itemize]{leftmargin=6.0mm}

\newcommand\benchmark{\makebox{ProtGlycanDock}}
\definecolor{r1}{RGB}{1,121,111}
\definecolor{r2}{RGB}{41,190,135}
\definecolor{r3}{RGB}{152,251,152}
\definecolor{tbd}{RGB}{255,0,0}

\definecolor{failcase}{RGB}{102,102,102}

\title{{\benchmark}: Towards Accurate \\ Protein-Glycan Docking with \\ Tailored Dataset, Benchmark and Model}

\author{
	\textbf{Minghao Xu}\textsuperscript{\rm 1$\,*$} \quad
	\textbf{Jiaze Song}\textsuperscript{\rm 1$\,*$} \quad
	\textbf{Jinyu Wang}\textsuperscript{\rm 2} \quad
	\textbf{Wentao Zhang}\textsuperscript{\rm 1,3,4$\,\dagger$} \\
	\textsuperscript{\rm *}{\small equal contribution} \quad
	\textsuperscript{\rm $\dagger$}{\small corresponding author} \\
	\textsuperscript{\rm 1}Peking University \quad
	\textsuperscript{\rm 2}University of Chinese Academy of Sciences \quad
	\textsuperscript{\rm 3}Zhongguancun Academy \\
	\textsuperscript{\rm 4}Beijing Key Laboratory of Data Intelligence and Security (Peking University) \\
	{\small \textbf{contacts:} minghao.xu@stu.pku.edu.cn, $\;$wentao.zhang@pku.edu.cn}
}

\begin{document}

\maketitle



\begin{abstract}

Performing accurate protein-glycan docking is critical to understanding protein-glycan interactions which govern many biological processes. However, this important task is hampered by the lack of tailored datasets, benchmarks and models for protein-glycan docking. To fill this blank, we first curate a standard benchmark \textbf{{\benchmark}} for training and evaluating protein-glycan docking models under both protein and glycan generalization settings. On this benchmark, we evaluate ten models in three categories, physics-based docking methods, machine-learning-based docking models and molecular structure prediction models. To obtain more precise binding poses, we further propose two finetuning techniques to inject the knowledge of glycan conformations into a general-purpose molecular structure prediction model. These two techniques respectively augment model inputs with glycan topological features and refine glycan binding pose with a glycan energy loss. Benchmark results show the superiority of AlphaFold~3 among existing methods, and the two proposed finetuning techniques can be well combined to achieve a state-of-the-art model for protein-glycan docking. Also, we perform stratified benchmark analysis and case studies to better understand the strengths and weaknesses of current models, providing insights for future model improvements. We maintain datasets at \url{https://huggingface.co/datasets/ProtGlycanDock/ProtGlycanDock} and source codes at \url{https://github.com/kasawa1234/ProtGlycanDock}. 

\end{abstract}



\section{Introduction} \label{sec:intro}

Protein-glycan interactions play pivotal roles in a wide spectrum of biological processes, including cell metabolism~\citep{koropatkin2012glycan}, immune response~\citep{van2008protein}, pathogen recognition~\citep{davicino2011coupling} and cellular adhesion~\citep{elola2007galectins}. Therefore, it is of great interest to solve the 3D structures of protein-glycan complexes, which lays the foundation for studying their interactions. However, the structural characterization of protein-glycan complexes remains experimentally challenging due to the inherent flexibility, branching diversity and weak binding affinity often exhibited by glycans~\citep{gimeno2020glycan}. Computational molecular docking offers a powerful complementary approach to predict protein-glycan complex structures~\citep{nieto2024computational}. 

Existing molecular docking~\citep{dominguez2003haddock,trott2010autodock,corso2022diffdock,cao2025surfdock} and structure prediction~\citep{abramson2024accurate,krishna2024generalized,bytedance2025protenix,chai2024chai,passaro2025boltz} approaches are primarily designed to model protein-small molecule and protein-protein interactions. However, they give limited consideration to the unique physicochemical properties of glycans, such as their inherent flexibility and branching patterns. Moreover, existing protein-glycan complex benchmarks~\citep{nivedha2016vina,ranaudo2024modeling} typically contain less than 100 samples and do not support model training.  

To address this limitation, this work proposes the \textbf{{\benchmark}} benchmark supporting model training and evaluation for protein-glycan docking. To build the benchmark dataset, we collect high-quality protein-glycan complex structures stored in the Protein Data Bank (PDB)~\citep{berman2002protein}, and we further use PDBFixer~\citep{eastman2013openmm} and MODELLER~\citep{webb2016comparative} to repair incomplete structures and convert non-standard amino acid residues to their canonical counterparts. The benchmark dataset is split by holding out novel proteins and glycans respectively, thereby comprehensively assessing model performance under both protein and glycan generalization settings. 
On the benchmark, we evaluate a widely-used physics-based docking method, AutoDock Vina~\citep{trott2010autodock}, its variant specific to protein-glycan docking, Vina-Carb~\citep{nivedha2016vina}, two machine-learning-based docking models, DiffDock~\citep{corso2022diffdock} and SurfDock~\citep{cao2025surfdock}, and six general-purpose molecular structure prediction models, AlphaFold~3~\citep{abramson2024accurate}, RoseTTAFold All-Atom~\citep{krishna2024generalized}, Chai-1~\citep{chai2024chai}, Boltz-2~\citep{passaro2025boltz}, Protenix~\citep{bytedance2025protenix} and Protenix with structural constraints (\emph{i.e.}, Protenix-constraint)~\citep{bytedance2025protenix}. 

To predict more precise protein-glycan complex structures,  
we finetune the Protenix and Protenix-constraint (two models with open-source training pipeline) on the training set of the {\benchmark} benchmark. During finetuning, we endow these general-purpose models with the knowledge of glycan structures from two aspects, \emph{i.e.}, input representations and training signals. For input representations, we augment the model with the global branching topology of glycans by encoding their 2D backbone structures and feeding into the model through an embedder module. For training signals, to ensure the physical plausibility of predicted glycan conformations, we incorporate a loss term based on the Carbohydrate Intrinsic (CHI) energy function~\citep{nivedha2014importance}, which is minimized along with the original structure prediction losses. 

Benchmark results show the superior performance of AlphaFold~3 against other baseline models on protein-glycan docking. By using the proposed finetuning methods, the finetuned Protenix-constraint outperforms AlphaFold~3 with a clear margin. Ablation studies verify that the injection of glycan structural knowledge into input representation and training signal are two complementary finetuning techniques, where they can be well combined to achieve top-performing models. We also conduct stratified benchmark analysis and case studies to analyze the strengths and limitations of current models, so as to inspire future works on protein-glycan docking. 



\section{Related Work} \label{sec:rela}

\textbf{Molecular docking and structure prediction.} To get the complex structure of multiple molecules, two kinds of methods are commonly used, molecular docking and molecular structure prediction. Physics-based docking methods~\citep{dominguez2003haddock,trott2010autodock,nivedha2016vina} predict protein-ligand interactions through an effective scoring function and search algorithm, while modern machine-learning-based docking models~\citep{corso2022diffdock,cao2025surfdock} learn the spatial transformation of a ligand relative to a protein receptor. 

In a separate line of research, AlphaFold 3~\citep{abramson2024accurate} and RoseTTAFold All-Atom~\citep{krishna2024generalized} first achieved generalized molecular structure prediction for various biomolecules (proteins, DNAs, RNAs, small molecules, glycans and ions) and their complexes. Following their efforts, the later works sought to incorporate experimental structural constraints~\citep{chai2024chai,bytedance2025protenix} and predict binding affinity along with structures~\citep{passaro2025boltz}. However, these models lack tailored design to capture the characteristics of glycan structures, and thus are suboptimal for protein-glycan docking. In this work, we mitigate this defect by customized model finetuning for protein-glycan docking. 


\textbf{Glycan machine learning.} With the rapid growth of glycomics data~\citep{bojar2022glycoinformatics,li2022artificial}, machine learning shows great promise in predicting glycan functions. Early works used traditional machine learning approaches like support vector machines (SVMs) to learn patterns from mass spectrometry data~\citep{kumozaki2015machine}, identify glycosylation sites on a protein~\citep{caragea2007glycosylation,li2015glycomine} and perform glycan clustering~\citep{yamanishi2007glycan}. Recent works employed sequence-based models~\citep{taherzadeh2019sprint,bojar2020sweetorigins,bojar2020using,thomes2021glycowork} and graph neural networks (GNNs)~\citep{dai2021attention,alkuhlani2023gnngly,xu2025glycanml} to encode the 2D backbone structure of a glycan and further predict its functions, including species origin, immunogenicity, type of glycosylation and binding affinity with a protein. GlycanAA~\citep{xu2025modeling} captured all-atom structures of a glycan with a hierarchical GNN, and it outperformed previous methods considering only glycan backbone structures on various function prediction tasks. 

However, existing works mainly focused on predicting glycan functions while rarely studied glycan structure prediction, especially predicting protein-glycan docked structures which determine many critical biological processes like metabolism and immune response. In this work, we study protein-glycan docking in depth from the aspects of dataset, benchmark and model. 



\section{The {\benchmark} Benchmark} \label{sec:benchmark}

In this section, we first give an overview of the structural characteristics of glycans and protein-glycan complexes (Section~\ref{sec:benchmark:structure}), and next thoroughly introduce the procedure of building the benchmark dataset (Section~\ref{sec:benchmark:dataset}). 



\begin{wrapfigure}{r}{0.53\textwidth}
  \centering
  \vspace{-13pt}
  \label{fig2}
  \includegraphics[width=0.53\textwidth]{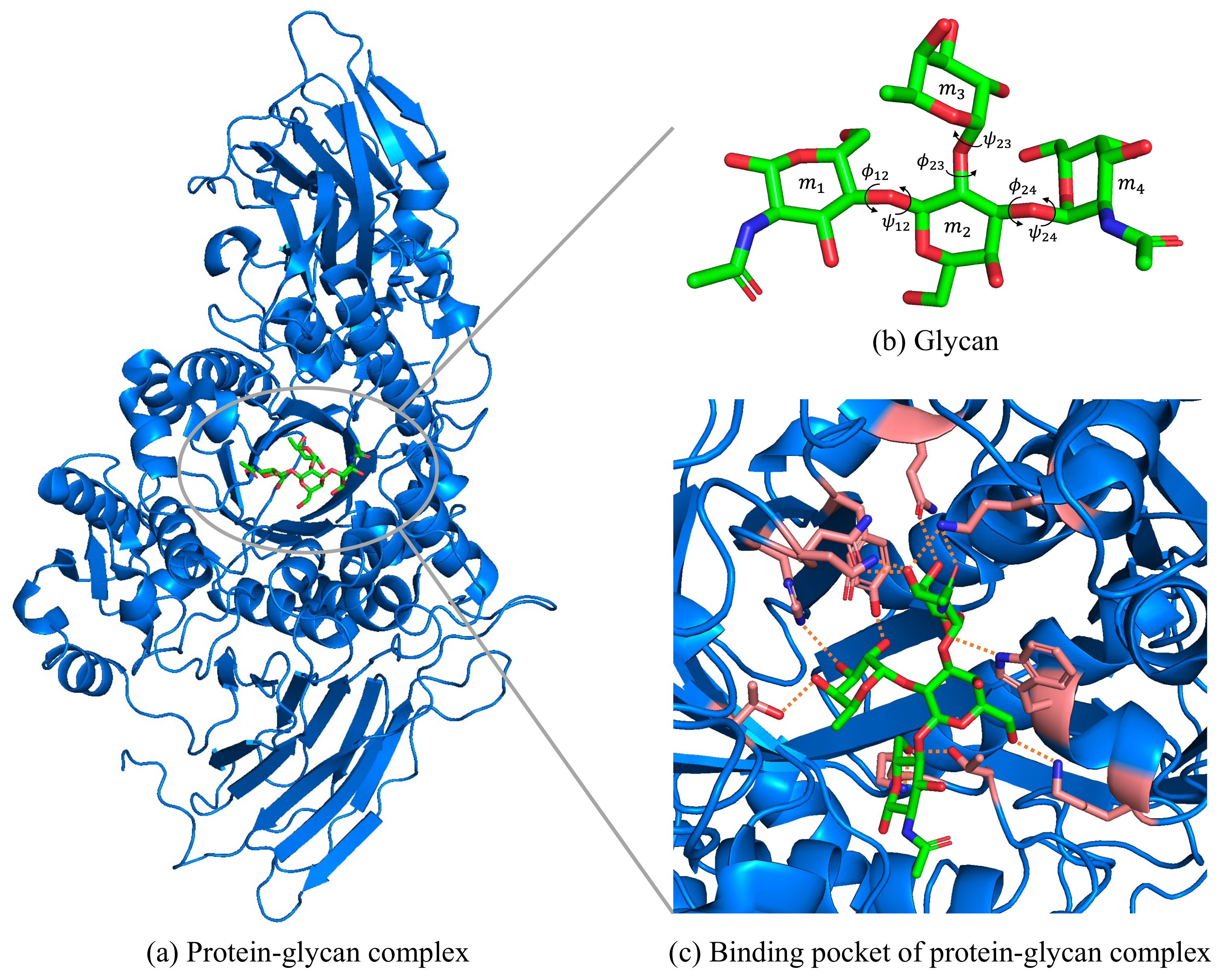}
  \vspace{-13.5pt}
  \caption{Structures of (a) a protein-glycan complex (PDB ID: 7Q1W), (b) the glycan in the complex, and (c) the binding pocket of the complex. \textcolor{orange}{Orange dashed lines} denote hydrogen bonds between glycan and binding pocket amino acid residues.}
  \label{fig:structure}
  \vspace{-10.5pt}
\end{wrapfigure}


\subsection{Navigating Glycan Structures and Their Complexes with Proteins} \label{sec:benchmark:structure}

In Figure~\ref{fig:structure}, we display a typical protein-glycan complex in the PDB database, an endo-beta-1,4-galactosidase in complex with blood group A tetrasaccharide (PDB ID: 7Q1W)~\citep{wu2021human}. A glycan is composed of multiple monosaccharides $\{m_i\}_{i=1}^N$ as its basic units, like the N-acetylglucosamine $m_1$, the galactose $m_2$, the fucose $m_3$ and the N-acetylgalactosamine $m_4$ in the glycan of Figure~\ref{fig:structure}(b). Different monosaccharides are linked by glycosidic bonds to form a complete glycan structure. Each glycosidic bond can be represented by two dihedral angles $\phi_{ij}$ and $\psi_{ij}$ (Figure~\ref{fig:structure}(b)), and the conformation of a glycan is primarily determined by the combination of all these dihedral angles on glycosidic bonds. 

When binding with a protein, a glycan interacts with the amino acid residues in the binding pocket of the protein, as shown in Figure~\ref{fig:structure}(c). Such interactions are driven by various forces, such as hydrogen bonding, hydrophobic interactions and electrostatic interactions. Among them, hydrogen bonding is the primary determinant for glycan positioning in the binding pocket, where hydrogen donor-acceptor pairs between glycan atoms and protein side chain atoms are attracted directionally to lock the glycan into a specific orientation (Figure~\ref{fig:structure}(c)). The binding pocket along with other protein domains form the complete protein-glycan complex (Figure~\ref{fig:structure}(a)). 


\subsection{Building the Benchmark Dataset} \label{sec:benchmark:dataset}

\textbf{Data collection.} We first extract all the structures with at least one glycan entity from the PDB (CC0 1.0 license), for a total of 13,716 structures. Protein-glycan interactions within these structures can be distinguished as two types, \emph{i.e.}, covalent glycosylation and noncovalent binding. Glycosylation is a covalent modification of protein surface residues, which is out of the scope of protein-glycan binding studied in this work. Therefore, we keep only 2,150 noncovalent binding protein-glycan complex structures. To prevent some overly complicated structures from interfering with overall benchmark results, we remove the structures with more than 4 protein chains, resulting in the final 1,781 structures. In Appendix~\ref{sec:app:statistics}, we provide the statistics of proteins and glycans in these structures. 

\textbf{Data fixing.} By navigating the raw PDB structures, we find two typical problems hindering structure modeling: (1) the existence of non-standard amino acid residues, and (2) the incomplete protein structures. To address the first problem, we employ the PDBFixer~\citep{eastman2013openmm} to substitute all non-standard amino acid residues occurred in the dataset with their structurally most similar counterpart within 20 common residues, followed by local structure relaxation through Amber energy minimization~\citep{hornak2006comparison}. To repair incomplete protein structures, we perform targeted loop modeling with MODELLER~\citep{webb2016comparative}, generating and selecting the best model based on the Discrete Optimized Protein Energy (DOPE) score to complement any missing residues. Notably, for all structures, the residues requiring fixing are beyond 6\AA$\,$ away from the glycan, thereby leaving binding pockets fully unaffected. All 1,781 collected structures successfully pass these two fixing steps, forming the final dataset. 

\textbf{Dataset splitting.} To comprehensively evaluate how well a model generalizes across unseen glycans and proteins, we build dedicated glycan-holdout and protein-holdout test sets respectively as below:
\begin{itemize}
    \item \emph{Glycan-holdout test set}: The whole dataset contains 332 distinct glycans. After sorting them by their earliest PDB deposition date, we hold out the 45 latest deposited glycans and their 134 corresponding protein–glycan samples to form the glycan-holdout test set. This test set evaluates the model’s generalizability to novel glycans identified in cutting-edge biological research. 
    \item \emph{Protein-holdout test set}: Given that the newest protein-glycan complexes have already been largely assigned to the glycan-holdout test set, and the remaining post-cutoff PDB entries (the latest training cutoff of baseline models: 2023-06-01) have close homologs or even the same proteins in earlier deposits, we use MMseqs2~\citep{steinegger2017mmseqs2} cluster-based split to avoid baseline memorization to the greatest extent. After clustering, we hold out 31 under-represented protein clusters (with 148 protein-glycan samples) for test, where 20 of the clusters have fewer than 5 PDB samples and the largest have only 8, making the evaluated proteins both dissimilar to our benchmark's training data and hard to memorize from PDB-scale pre-training.  
\end{itemize}
For the rest 280 protein clusters, we count the number of glycans that only occur in the protein-glycan samples of each cluster, and the 25 clusters with most such within-cluster unique glycans are selected. The 125 protein-glycan samples in these 25 clusters serve as the validation set. This validation set can evaluate model generalization to unseen glycans and proteins simultaneously, thus satisfying the criteria for model selection. Finally, the {\benchmark} benchmark is composed of 1,374 training, 125 validation, 134 glycan-holdout test, and 148 protein-holdout test samples. 



\section{Customized Finetuning for Protein-Glycan Docking} \label{sec:method}

\begin{figure*}[t]
\centering
    \includegraphics[width=1.0\linewidth]{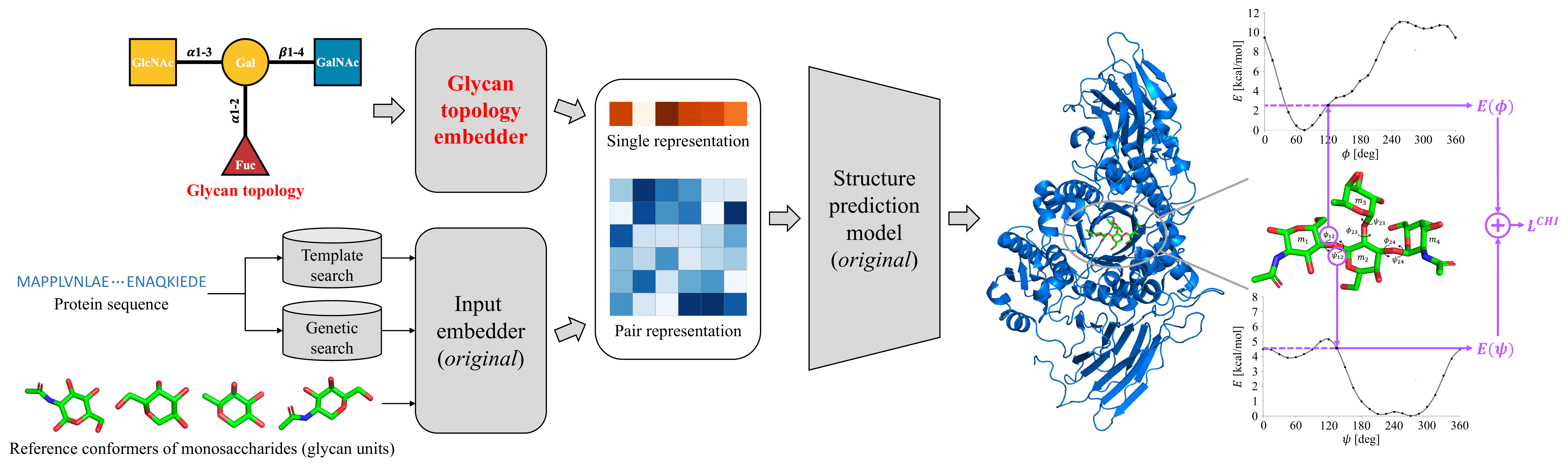}
    \vspace{-4.6mm}
    \caption{Illustration of finetuning a molecular structure prediction model with glycan topology embedding (\textcolor{red}{red part}) and Carbohydrate Intrinsic (CHI) energy loss ({\color[rgb]{0.7695,0.4258,0.9961} purple part}).}
    \label{fig:finetune}
\end{figure*}

Given a pre-trained molecular structure prediction model, we aim to enhance its ability to predict protein-glycan complex structures. To achieve this, we finetune the model by incorporating glycan topological features into its input (Section~\ref{sec:method:input}) and employing a training loss that focuses on glycosidic bond rotation (Section~\ref{sec:method:loss}), as shown in Figure~\ref{fig:finetune}. 


\subsection{Input Representation of Glycan Topology} \label{sec:method:input}

Existing molecular structure prediction models~\citep{abramson2024accurate,chai2024chai,passaro2025boltz,bytedance2025protenix} commonly deem a glycan as a group of connected small molecules (each small molecule denotes a monosaccharide in the glycan), and they represent each small molecule by its reference conformer, as illustrated in the lower left corner of Figure~\ref{fig:finetune}. By using the features of reference conformers as input, the model can well capture the local atomic structure of each monosaccharide in a glycan. However, this method cannot capture the global topology of a glycan which defines the glycan's overall backbone structure. To tackle this limitation, we inject the features of glycan topology into the model through single and pair representations, fitting the original design of mainstream models~\citep{abramson2024accurate,chai2024chai,passaro2025boltz,bytedance2025protenix}: 
\begin{itemize}
\vspace{-2.8mm}
    \item \textbf{Single representation of glycan topology}: In single representation, we would like to incorporate the type information of different monosaccharides in a glycan topology. Specifically, we represent the type of each monosaccharide with a 37-dimensional one-hot vector and map it to the space of single representation with an embedding layer, where 37 types of monosaccharides occurred in the PDB database are considered. The embedding of a monosaccharide is added to the single representations of all its associated atoms. 
    \vspace{-0.6mm}
    \item \textbf{Pair representation of glycan topology}: In pair representation, we would like to embed the graphical structure of a glycan formed by glycosidic bonds. In specific, we represent each glycosidic bond with an 11-dimensional one-hot vector and use an embedding layer to map the vector to the space of pair representation, where we consider 11 types of glycosidic bonds occurred in the PDB. The embedding of a glycosidic bond is added to its associated atom pair, and the pair representations of the atom pairs without glycosidic linkage are not modified. 
\vspace{-2.5mm}
\end{itemize}
The single and pair representations after glycan topology injection are fed into the model, and the model is finetuned to adapt these changed inputs. 


\subsection{Training Loss for Glycosidic Bond Geometry} \label{sec:method:loss}

When binding with a protein, a glycan tends to adopt a conformation with low energy~\citep{gimeno2020glycan}. As stated in Section~\ref{sec:benchmark:structure}, the dihedral angles $\phi$ and $\psi$ of glycosidic bonds mainly determine the conformation of a glycan and further determine its energy. Therefore, in a predicted protein-glycan complex, we aim to minimize the glycan energy with respect to the dihedral angles of glycosidic bonds. 

To achieve this, we adopt the Carbohydrate Intrinsic (CHI) energy function~\citep{nivedha2014importance} to represent glycan energy with glycosidic dihedral angles. The CHI energy function approximates quantum mechanical rotational energies with Gaussian expansions, and the energy functions of both $\phi$ and $\psi$ are respectively calculated for $\alpha$- and $\beta$-linkages (two stereoisomeric configurations of glycosidic bonds): 
\begin{fleqn}[0pt]
\begin{equation} \label{eq:phi_alpha}
\fontsize{8.6}{8}\selectfont
\begin{split}
& \mathbf{\phi} \textbf{ in } \mathbf{\alpha}\textbf{-linkages:}\\
& E(\phi) = 2.977e^{-\frac{(\phi+199.49)^2}{677.81}} + 102.25e^{-\frac{(\phi-170.6)^2}{1696.8}} + 10.745e^{-\frac{(\phi+105.31)^2}{4724.6}} + 3.674e^{-\frac{(\phi-6.201)^2}{1347.7}} + \\ 
& \qquad\quad\;\;\; 2.061e^{-\frac{(\phi-91.655)^2}{1500}} + 6.194e^{-\frac{(\phi+22.979)^2}{2122.3}} - 2.112e^{-\frac{(\phi-83.602)^2}{1254.1}} - 98.001e^{-\frac{(\phi-170.01)^2}{1598.7}} ,
\end{split}
\end{equation}
\end{fleqn}
\begin{fleqn}[0pt]
\begin{equation} \label{eq:phi_beta}
\fontsize{8.6}{8}\selectfont
\begin{split}
& \mathbf{\phi} \textbf{ in } \mathbf{\beta}\textbf{-linkages:} \\
& E(\phi) = 450.54e^{-\frac{(\phi+330.77)^2}{4449.8}} + 23.712e^{-\frac{(\phi-304.63)^2}{8375.2}} + 5.935e^{-\frac{(\phi+152.08)^2}{6049.8}} + 22.467e^{-\frac{(\phi+23.516)^2}{606.9}} + \\
& \qquad\quad\;\;\; 10.036e^{-\frac{(\phi-120.96)^2}{4038}} - 18.141e^{-\frac{(\phi+24.268)^2}{543.05}} + 5.882e^{-\frac{(\phi-19.632)^2}{897.93}} - 2.128 ,
\end{split}
\end{equation}
\end{fleqn}
\begin{fleqn}[0pt]
\begin{equation} \label{eq:psi_alpha}
\fontsize{8.6}{8}\selectfont
\begin{split}
& \mathbf{\psi} \textbf{ in } \mathbf{\alpha}\textbf{-linkages:} \\
& E(\psi) = 4.624e^{-\frac{(\psi-5.046)^2}{5005.8}} + 4.614e^{-\frac{(\psi-362.49)^2}{2090.6}} + 4.942e^{-\frac{(\psi-121.2)^2}{2093.8}} + 0.403e^{-\frac{(\psi-241.43)^2}{456.83}} + \\
& \qquad\quad\;\;\; 0.799e^{-\frac{(\psi-68.425)^2}{678.81}} + 0.223e^{-\frac{(\psi-192.93)^2}{347.25}} - 0.126 ,
\end{split}
\end{equation}
\end{fleqn}
\begin{fleqn}[0pt]
\begin{equation} \label{eq:psi_beta}
\fontsize{8.6}{8}\selectfont
\begin{split}
& \mathbf{\psi} \textbf{ in } \mathbf{\beta}\textbf{-linkages:} \\
& E(\psi) = 4.468e^{-\frac{(\psi-10^{-30})^2}{1279.6}} + 4.382e^{-\frac{(\psi-357.77)^2}{6050.1}} + 284.95e^{-\frac{(\psi-146.64)^2}{1551.8}} + 4.761e^{-\frac{(\psi-220.68)^2}{5892.9}} - \\
& \qquad\quad\;\;\; 169.2e^{-\frac{(\psi-147.37)^2}{1742.5}} - 118.44e^{-\frac{(\psi-146.06)^2}{1359.8}} + 1.022 ,
\end{split}
\end{equation}
\end{fleqn}
where $-180^{\circ} < \phi < 180^{\circ}$ and $0^{\circ} < \psi < 360^{\circ}$. The CHI energy loss $L^{\mathrm{CHI}}$ is defined as the average of CHI energies on all $K$ glycosidic bonds in a glycan: 
\begin{equation} \label{eq:chi_loss}
    L^{\mathrm{CHI}} = \frac{1}{2K} \sum_{k=1}^{K} \big( E(\phi_k) + E(\psi_k) \big) . 
\end{equation}
The CHI energy loss is minimized along with the original structure prediction loss $L^{\mathrm{pred}}$ as below: 
\begin{equation} \label{eq:total_loss}
    L = L^{\mathrm{pred}} + \gamma \;\! L^{\mathrm{CHI}} ,
\end{equation}
where $\gamma$ is the weight for CHI energy loss, and $L$ denotes the total loss for model finetuning. In this way, the model learns to refine the glycan conformation in a predicted protein-glycan complex.  


\section{Experiments} \label{sec:exp}

\subsection{Experimental Setups} \label{sec:exp:setup}

\textbf{Baseline models.} We adopt two physics docking methods, AutoDock Vina~\citep{trott2010autodock} and Vina-Carb~\citep{nivedha2016vina}, two machine-learning docking models, DiffDock~\citep{corso2022diffdock} and SurfDock~\citep{cao2025surfdock}, and six machine-learning molecular structure prediction models, RoseTTAFold All-Atom~\citep{krishna2024generalized}, Chai-1~\citep{chai2024chai}, Boltz-2~\citep{passaro2025boltz}, AlphaFold 3~\citep{abramson2024accurate}, Protenix~\citep{bytedance2025protenix} and Protenix-constraint~\citep{bytedance2025protenix}. All models are evaluated under the blind-docking setting (no prior ground-truth protein receptor structure and binding site information) and inferred with seed 42, 43, 44, 45 and 46 to report error bars. The detailed setups of each model are stated in Appendix~\ref{sec:app:baseline}. 

\textbf{Finetuned models.} Besides the vanilla finetuning, we finetune the Protenix and Protenix-constraint with three proposed settings: 
\begin{itemize}
\vspace{-2.3mm}
    \item \textbf{+ Glycan Topology Embedding}: Only the model input is augmented by the proposed glycan topology embedding scheme. 
    \vspace{-0.6mm}
    \item \textbf{+ CHI Loss}: Only the loss function incorporates the proposed CHI energy loss with the loss weight $\gamma=0.01$. The choice of loss weight is based on the scale of CHI loss against original structure prediction loss. We give a sensitivity analysis on $\gamma$ in Appendix~\ref{sec:app:sensitivity}. 
    \vspace{-0.6mm}
    \item \textbf{+ Glycan Topology Embedding \& CHI Loss}: Both model input and loss function are augmented as above. 
\vspace{-2mm}
\end{itemize}

\textbf{Finetuning setups.} Our finetuning configuration is based on the Finetuning Stage 2 setting provided by Protenix (training crop size: 640; with training the PAE head, distogram head and diffusion module). We set the EMA decay rate to 0.9999 and the learning rate to $1 \times 10^{-4}$. Under all settings, the model is finetuned for 15,000 steps with an Adam optimizer. For model selection, validation is performed every 100 training steps, and the checkpoint achieving the highest average ranking score over all validation samples is selected for test. All experiments are conducted on a local server equipped with 100 CPU cores and 8 NVIDIA HGX H20 GPUs (96 GB). 



\textbf{Evaluation metrics.} Following previous practices~\citep{trott2010autodock,abramson2024accurate,bytedance2025protenix,cao2025surfdock}, we evaluate the performance of protein-glycan docking from two aspects, geometric correctness and physical plausibility. We adopt Ligand RMSD, Ring sRMSD, Complex RMSD and protein-glycan contact recovery rate to assess the geometric correctness of predicted protein-glycan complexes, report the number of predictions satisfying Ligand RMSD $<2$\AA$\,$ as geometrically correct samples, and report the number of predictions both satisfying Ligand RMSD $<2$\AA$\,$ and passing all PoseBusters tests~\citep{buttenschoen2024posebusters} as the samples that are both geometrically correct and physically plausible. These metrics are defined as below:
\begin{itemize}
\vspace{-2mm}
    \item \textbf{Ligand RMSD}: After superposing the predicted protein onto the ground-truth protein, the Root-Mean-Square Deviation (RMSD) between the predicted glycan and the ground-truth glycan is computed as Ligand RMSD. 
    \vspace{-0.6mm}
    \item \textbf{Ring sRMSD}: This metric specifically evaluates glycan conformation accuracy. It aligns all ring atoms of predicted glycan onto their counterparts in ground truth, and the alignment RMSD is defined as Ring sRMSD. 
    \vspace{-0.6mm}
    \item \textbf{Complex RMSD}: The Complex RMSD is computed over all atoms after a global superposition of the predicted complex structure onto the ground-truth complex structure. 
    \item \textbf{Contact recovery rate}: It calculates the ratio of the contacting residue-monosaccharide (glycan unit) pairs recovered in the predicted complex structure. 
    \item \textbf{PoseBusters tests}: These tests evaluate ligand stereochemistry, bond lengths and angles, aromatic ring and double-bond planarity, internal steric clashes, and conformational energy. They also check for steric clashes and excessive volume overlap between the ligand and the protein in a complex. 
\end{itemize}

\begin{table*}[t]
\centering
\caption{Benchmark results on {\benchmark}. Mean (standard deviation) of each metric is given. The best, second-best, and third-best results are denoted by \textbf{bold}, \underline{underline}, and \textit{italic}, respectively. \emph{Abbr.}, PB-valid: passing all PoseBusters tests, FT.: Finetuned; Topo.: Topology; Emb.: Embedding.} 
\label{tab:benchmark}
\begin{spacing}{1.3}
\vspace{-0.6mm}
\begin{adjustbox}{max width=1\linewidth}
\begin{tabular}{l|cccccc|cccccc}
\toprule
\multirow{4}{*}{\bf{\large Model}} & \multicolumn{6}{c|}{\bf{\large Protein-holdout test (148 samples)}} & \multicolumn{6}{c}{\bf{\large Glycan-holdout test (134 samples)}} \\
\cmidrule{2-13}
& \multirow{3}{*}{ \makecell[c]{\bf{\#Ligand} \\ \bf{RMSD $<2$\AA $\uparrow$}}} &
\multirow{3}{*}{ \makecell[c]{\bf{\#Ligand} \\ \bf{RMSD $<2$\AA} \\ \bf{\& PB-valid $\uparrow$}}} & 
\multirow{3}{*}{ \makecell[c]{\bf{Ligand} \\ \bf{RMSD $\downarrow$}}} &
\multirow{3}{*}{ \makecell[c]{\bf{Ring} \\ \bf{sRMSD $\downarrow$}}} &
\multirow{3}{*}{ \makecell[c]{\bf{Complex} \\ \bf{RMSD $\downarrow$}}} &
\multirow{3}{*}{ \makecell[c]{\bf{Contact} \\ \bf{Recovery\% $\uparrow$}}} &
\multirow{3}{*}{ \makecell[c]{\bf{\#Ligand} \\ \bf{RMSD $<2$\AA $\uparrow$}}} &
\multirow{3}{*}{ \makecell[c]{\bf{\#Ligand} \\ \bf{RMSD $<2$\AA} \\ \bf{\& PB-valid $\uparrow$}}} & 
\multirow{3}{*}{ \makecell[c]{\bf{Ligand} \\ \bf{RMSD $\downarrow$}}} &
\multirow{3}{*}{ \makecell[c]{\bf{Ring} \\ \bf{sRMSD $\downarrow$}}} &
\multirow{3}{*}{ \makecell[c]{\bf{Complex} \\ \bf{RMSD $\downarrow$}}} &
\multirow{3}{*}{ \makecell[c]{\bf{Contact} \\ \bf{Recovery\% $\uparrow$}}} \\
&  &  &  &  &  &  &  &  &  &  &  &  \\
&  &  &  &  &  &  &  &  &  &  &  &  \\
\midrule
\multicolumn{13}{c}{\bf{\large Physics-based docking methods}} \\
\midrule
AutoDock Vina & 1.2 (0.4) & 0.6 (0.5) & 23.05 (0.62) & 2.60 (0.11) & 13.87 (0.51) & 17.2 (0.6) & 1.8 (0.4) & 1.6 (0.8) & 24.47 (0.94) & 1.98 (0.17) & 7.63 (0.22) & 17.5 (0.5) \\
Vina-Carb & 2.0 (0.6)  & 1.2 (0.4) & 22.48 (0.53) & 2.23 (0.13) & 11.65 (0.31) & 18.7 (0.8) & 3.8 (0.7) & 2.6 (0.5) & 22.95 (0.87) & 1.71 (0.23) & 7.10 (0.29) & 19.3 (1.1) \\
\midrule
\multicolumn{13}{c}{\bf{\large Machine-learning-based docking models}} \\
\midrule
DiffDock & 12.8 (0.4) & 8.4 (0.5) & 17.22 (0.45) & 1.68 (0.10) & 9.17 (0.24) & 24.9 (1.2) & 12.6 (0.8) & 8.2 (0.4) & 16.95 (0.49) & 1.49 (0.13) & 5.24 (0.22) & 25.8 (0.9) \\
SurfDock & 15.2 (0.7) & 10.6 (0.8) & 16.31 (0.37) & 1.42 (0.11) & 8.55 (0.21) & 26.3 (0.7) & 14.2 (0.4) & 9.4 (0.5) & 15.78 (0.42) & 1.42 (0.07) & 4.61 (0.19) & 27.1 (0.7) \\
\midrule
\multicolumn{13}{c}{\bf{\large Molecular structure prediction models}} \\
\midrule
RoseTTAFold All-Atom  & 1.8 (0.4) & 1.4 (0.5) & 20.78 (0.47) & 2.97 (0.19) & 12.75 (0.34) & 19.3 (0.8) & 3.6 (0.5) & 3.0 (0.9) & 23.36 (0.42) & 1.94 (0.13) & 7.09 (0.34) & 20.5 (0.7) \\
Chai-1              & 20.6 (1.0) & 14.4 (0.5) & 15.67 (0.42) & 1.10 (0.12) & 9.53 (0.25) & 28.8 (0.7) & 24.2 (0.7) & 16.8 (0.4) & 15.71 (0.38) & 1.32 (0.13) & 4.72 (0.26) & 29.3 (0.5) \\
Boltz-2              & 19.6 (0.8) & 13.6 (0.5) & 15.76 (0.37) & 1.62 (0.13) & 8.94 (0.19) & 27.2 (0.6) & 25.0 (0.6) & 16.2 (0.7) & 14.95 (0.41) & 1.27 (0.09) & 4.54 (0.24) & 27.7 (1.0) \\
AlphaFold 3          & 34.6 (0.5) & 24.8 (0.4) & \textit{14.27} (0.54) & \textbf{0.73} (0.05) & 7.98 (0.23) & 34.7 (1.4) & 30.6 (1.0) & 21.2 (0.7) & 13.87 (0.65) & 1.24 (0.11) & 3.76 (0.28) & 35.4 (1.1) \\
Protenix & 26.8 (0.7) & 18.2 (0.4) & 17.73 (0.29) & 1.12 (0.07) & 8.83 (0.22) & 30.6 (0.7) & 25.6 (0.5) & 19.2 (0.4) & 14.75 (0.33) & 1.27 (0.09) & 4.59 (0.21) & 31.1 (0.5) \\
Protenix-constraint & 30.0 (0.6) & 22.2 (0.7) & 17.54 (0.43) & 0.98 (0.06) & 9.17 (0.27) & 32.8 (1.2) & 28.4 (0.5) & 20.6 (0.5) & 14.63 (0.51) & 1.38 (0.13) & 4.45 (0.20) & 34.3 (1.2) \\
\midrule
\multicolumn{13}{c}{\bf{\large Finetuned molecular structure prediction models}} \\
\midrule
FT. Protenix
                    & 31.4 (0.5) & 21.8 (0.7) & 16.86 (0.34) & 0.95 (0.11) & 8.12 (0.35) & 33.4 (0.8) & 31.6 (1.0) & 22.0 (0.6) & 13.52 (0.44) & 1.22 (0.06) & 4.15 (0.21) & 33.8 (0.9) \\
\makecell[l]{FT. Protenix \\[-0.5mm] {\small \textsc{$\;$ + Topo. Emb.}}}
                    & 33.2 (0.4) & 23.2 (0.7) & 16.09 (0.37) & 0.95 (0.06) & 7.97 (0.26) & 34.8 (0.7) & 32.6 (0.5) & 23.0 (1.4) & 13.08 (0.30) & 1.20 (0.05) & 3.92 (0.24) & 34.3 (1.0) \\
\makecell[l]{FT. Protenix \\[-0.5mm] {\small \textsc{$\;$ + CHI Loss}}}
                    & 34.6 (0.5) & 24.0 (0.6) & 15.57 (0.28) & 0.93 (0.10) & 7.66 (0.23) & 35.9 (1.1) & 33.6 (1.0) & 23.2 (0.7) & 12.85 (0.31) & 1.17 (0.08) & 3.79 (0.26) & 34.5 (0.5) \\
\makecell[l]{FT. Protenix \\[-0.5mm] {\small \textsc{$\;$ + Topo. Emb. \& CHI Loss}}}
                    & 36.8 (0.7) & \textit{25.6} (0.5) & 15.12 (0.23) & \textit{0.85} (0.05) & \textit{7.50} (0.28) & 38.0 (0.9) & 34.8 (0.7) & \textit{24.4} (0.5) & 12.52 (0.27) & 1.15 (0.09)  & 3.73 (0.21) & 37.2 (1.1) \\

\cmidrule(lr){1-13}
FT. Protenix-constraint
                    & 35.6 (0.5) & 25.0 (0.6) & 15.32 (0.47) & 0.92 (0.09) & 7.90 (0.33) & 40.8 (1.2) & 32.4 (0.5) & 23.8 (0.4) & 12.76 (0.41) & 1.17 (0.07) & 3.83 (0.25) & 39.6 (1.2) \\
\makecell[l]{FT. Protenix-constraint \\[-0.5mm] {\small \textsc{$\;$ + Topo. Emb.}}}
                    & \textit{37.2} (0.7) & 24.8 (0.4) & \underline{13.96} (0.35) & 0.90 (0.07) & 8.02 (0.29) & \textit{41.7} (0.8) & \textit{35.8} (0.4) & 23.6 (0.8) & \textit{12.04} (0.30) & \textit{1.13} (0.06) & \textit{3.61} (0.19) & \textit{40.9} (1.0) \\
\makecell[l]{FT. Protenix-constraint \\[-0.5mm] {\small \textsc{$\;$ + CHI Loss}}}
                    & \underline{38.6} (0.5) & \underline{25.8} (0.4) & 14.61 (0.53) & 0.86 (0.05) & \underline{7.37} (0.26) & \underline{42.3} (0.7) & \underline{37.0} (0.6) & \underline{25.6} (0.8) & \underline{11.82} (0.33) & \underline{1.10} (0.05) & \underline{3.26} (0.16) & \underline{42.0} (0.8) \\
\makecell[l]{FT. Protenix-constraint \\[-0.5mm] {\small \textsc{$\;$ + Topo. Emb. \& CHI Loss}}}
                    & \textbf{42.0} (0.6) & \textbf{31.2} (0.4) & \textbf{11.93} (0.46) & \underline{0.84} (0.06) & \textbf{6.79} (0.25) & \textbf{44.7} (0.8) & \textbf{42.2} (0.7) & \textbf{31.8} (0.7) & \textbf{11.03} (0.57) & \textbf{1.02} (0.07) & \textbf{2.89} (0.20) & \textbf{45.2} (1.3) \\
\bottomrule
\end{tabular}
\end{adjustbox}
\end{spacing}
\vspace{-1.3mm}
\end{table*}




\subsection{Overall Benchmark Results} \label{sec:exp:result}
\vspace{-0.2mm}

In Table~\ref{tab:benchmark}, we report overall evaluation results on the protein-holdout and glycan-holdout test sets of the {\benchmark} benchmark. We report the mean and standard deviation of each metric over five inference seeds. 
According to these results, we highlight the following key findings:
\begin{itemize}
\vspace{-2mm}
    \item \textbf{Machine-learning-based docking is superior to physics-based docking in protein-glycan docking tasks.} DiffDock and SurfDock produces more geometrically valid and physically plausible docking samples than AutoDock Vina and Vina-Carb, and they also recover more residue-monosaccharide contacts on the protein-glycan interface. Thus, learning-based methods generalize better across the glycan ligands of varying sizes. 
    \vspace{-0.6mm}
    \item \textbf{Molecular structure prediction models significantly outperform docking models under the blind docking setting.} On both test sets, all evaluated molecular structure prediction models except RoseTTAFold All-Atom achieve clearly more geometrically valid and physically plausible samples than four docking models, with lower RMSD metrics, higher contact recovery rate, and more samples passing all PoseBusters tests. Therefore, molecular structure prediction models are better suited to blind protein-glycan docking tasks. 
    \vspace{-0.6mm}
    \item \textbf{AlphaFold 3 is the best-performing model without finetuning.} AlphaFold 3 outperforms other non-finetuned models in terms of RMSD metrics, contact recovery rate, and the number of geometrically valid and physically plausible samples. Especially, on the protein-holdout test set, AlphaFold 3 achieves the lowest Ring sRMSD among all models including finetuned ones, illustrating its superiority in glycan binding pose prediction. 
    \vspace{-0.6mm}
    \item \textbf{The protein-glycan docking capability of a model can be well enhanced by finetuning on protein-glycan complex data.} After finetuning on the training set of the {\benchmark} benchmark, Protenix and Protenix-constraint both produce more geometrically valid and physically plausible docking samples, and their RMSD metrics and contact recovery rate are also improved. This result verifies the effectiveness of domain-specific finetuning in boosting more accurate protein-glycan docking. 
    \vspace{-0.6mm}
    \item \textbf{Both glycan topology embedding and CHI loss can improve the effectiveness of finetuning, and they can be well combined to yield top-performing models.} For both Protenix and Protenix-constraint, when incorporating either glycan topology embedding or CHI loss during finetuning, we can observe performance improvements in RMSD metrics, contact recovery rate, and the number of geometrically valid and physically plausible docking samples on both test sets. By jointly using these two techniques, the finetuned Protenix-constraint achieves the best performance on 11 out of 12 evaluation metrics. These results demonstrate that the two proposed finetuning techniques are mutually complementary, and their integration delivers superior protein-glycan docking performance. 
\vspace{-1.5mm}
\end{itemize}

\subsection{Stratified Benchmark Analysis} \label{sec:exp:analysis}

In Table~\ref{tab:analysis}, we report the performance of six competitive models varied by glycan size, protein size and data deposition time. We stratify the results by glycan size on the glycan-holdout test set, by protein size on the protein-holdout test set, and by data deposition time on the protein-holdout test set. 
We analyze each stratification in the following parts. 

\textbf{Model performance varied by glycan size.} We observe performance decay across all six models as glycan size increases, illustrating their limitations in predicting protein–glycan complexes containing large glycans. The Protenix-constraint finetuned with glycan topology embedding and CHI loss achieves best performance on all 3 glycan-stratified metrics. Especially, it narrows the performance gap between small and large glycans, where, in terms of the Ligand RMSD gap between the samples with more than 7 monosaccharides and the ones with 2-4 monosaccharides, it achieves a gap of 9.02~\AA{} against the gap of 19.88~\AA{} achieved by the non-finetuned Protenix-constraint. Therefore, the proposed finetuning strategy can enhance model generalization to large glycans. 

\textbf{Model performance varied by protein size.} Performance decay also occurs across all six models as protein size increases, indicating their limitations in large-protein-involved protein-glycan docking. The Protenix-constraint finetuned with glycan topology embedding and CHI loss achieves the best results across all 3 protein-stratified metrics. Also, it reduces the performance discrepancy between small and large proteins. Specifically, when comparing the Ligand RMSD gap between the samples with more than 1000 amino acids and the ones with 64-500 amino acids, the best-performing finetuned Protenix-constraint yields a gap of 4.99~\AA{}, in contrast to the larger gap of 8.52~\AA{} obtained by its non-finetuned counterpart. These results demonstrate that the proposed finetuning strategy effectively strengthens the model’s generalization capability toward large proteins. 

\textbf{Model performance varied by data deposition time.} All six models exhibit a consistent, gradual increase in Ligand RMSD across these periods, indicating lower docking accuracy on more recently deposited structures. Nevertheless, the absolute differences remain modest, where the increase between the earliest and latest groups ranges from 1.38 to 1.80~\AA{}, remaining below 2~\AA{} for all models. The relative ranking of all six models is also unchanged across the three periods, suggesting that their comparative performance is consistent across temporal subsets. Overall, these models exhibit robust generalization to recently identified protein–glycan complex systems, with unchanged performance rankings across deposition periods. 


\begin{table*}[t]
\centering
\caption{The performance of six competitive models varied by glycan size, protein size and data deposition time. Mean (standard deviation) of each stratified Ligand RMSD metric is reported. The best result is denoted by \textbf{bold}. \emph{Abbr.}, Mono.: Monosaccharides; AA.: Amino Acids; FT.: Finetuned; Topo.: Topology; Emb.: Embedding.} 
\label{tab:analysis}
\begin{spacing}{1.3}
\vspace{-0.6mm}
\begin{adjustbox}{max width=1\linewidth}
\begin{tabular}{l|ccc|ccc|ccc}
\toprule
\multirow{3}{*}{\bf{Model}} & \multicolumn{3}{c|}{\bf{Glycan-stratified Ligand RMSD $\downarrow$}} & \multicolumn{3}{c|}{\bf{Protein-stratified Ligand RMSD $\downarrow$}} & \multicolumn{3}{c}{\bf{Time-stratified Ligand RMSD $\downarrow$}} \\
& \multicolumn{3}{c|}{\bf{on glycan-holdout test set}} & \multicolumn{3}{c|}{\bf{on protein-holdout test set}} & \multicolumn{3}{c}{\bf{on protein-holdout test set}} \\
\cmidrule{2-10}
& \bf{2-4 Mono.} & \bf{5-7 Mono.} & \bf{$>${7 Mono.}} & \bf{64-500 AA.} & \bf{501-1000 AA.} & \bf{$>$1000 AA.} & \bf{1990-2010} & \bf{2011-2020} & \bf{2021-2026} \\
\midrule
AlphaFold 3 & 11.45 (0.43) & 21.73 (0.57) & 31.92 (0.60) & 9.13 (0.48) & 13.18 (0.39) & 18.25 (0.45) & 12.17 (0.51) & 12.90 (0.38) & 13.97 (0.51) \\
Protenix-constraint & 12.18 (0.47) & 25.42 (0.63) & 32.06 (0.64) & 10.07 (0.35) & 13.95 (0.58) & 18.59 (0.43) & 13.40 (0.61) & 14.03 (0.46) & 14.78 (0.42) \\
FT. Protenix-constraint & 10.94 (0.38) & 21.13 (0.47) & 28.11 (0.43) & 8.87 (0.32) & 12.35 (0.40) & 15.98 (0.51) & 11.69 (0.46) & 12.48 (0.39) & 13.39 (0.54) \\
\makecell[l]{FT. Protenix-constraint \\[-0.5mm] {\small \textsc{$\;$ + Topo. Emb.}}}
                    & 10.63 (0.36) & 18.86 (0.48) & 24.95 (0.55) & 8.68 (0.33) & 11.76 (0.46) & 15.04 (0.42) & 11.10 (0.37) & 11.77 (0.41) & 12.68 (0.49) \\
\makecell[l]{FT. Protenix-constraint \\[-0.5mm] {\small \textsc{$\;$ + CHI Loss}}}
                    & 10.82 (0.38) & 16.78 (0.41) & 22.70 (0.47) & 8.73 (0.27) & 11.53 (0.35) & 14.70 (0.40) & 10.94 (0.32) & 11.59 (0.39) & 12.59 (0.42) \\
\makecell[l]{FT. Protenix-constraint \\[-0.5mm] {\small \textsc{$\;$ + Topo. Emb. \& CHI Loss}}}
                    & \textbf{10.00} (0.24) & \textbf{14.34} (0.31) & \textbf{19.02} (0.39) & \textbf{8.19} (0.23) & \textbf{10.57} (0.32) & \textbf{13.18} (0.39) & \textbf{9.89} (0.27) & \textbf{10.51} (0.31) & \textbf{11.55} (0.35) \\
\bottomrule
\end{tabular}
\end{adjustbox}
\end{spacing}
\vspace{-1.5mm}
\end{table*}




\subsection{Case Studies} \label{sec:exp:case}

\begin{figure*}[t]
    \centering
    \includegraphics[width=0.92\linewidth]{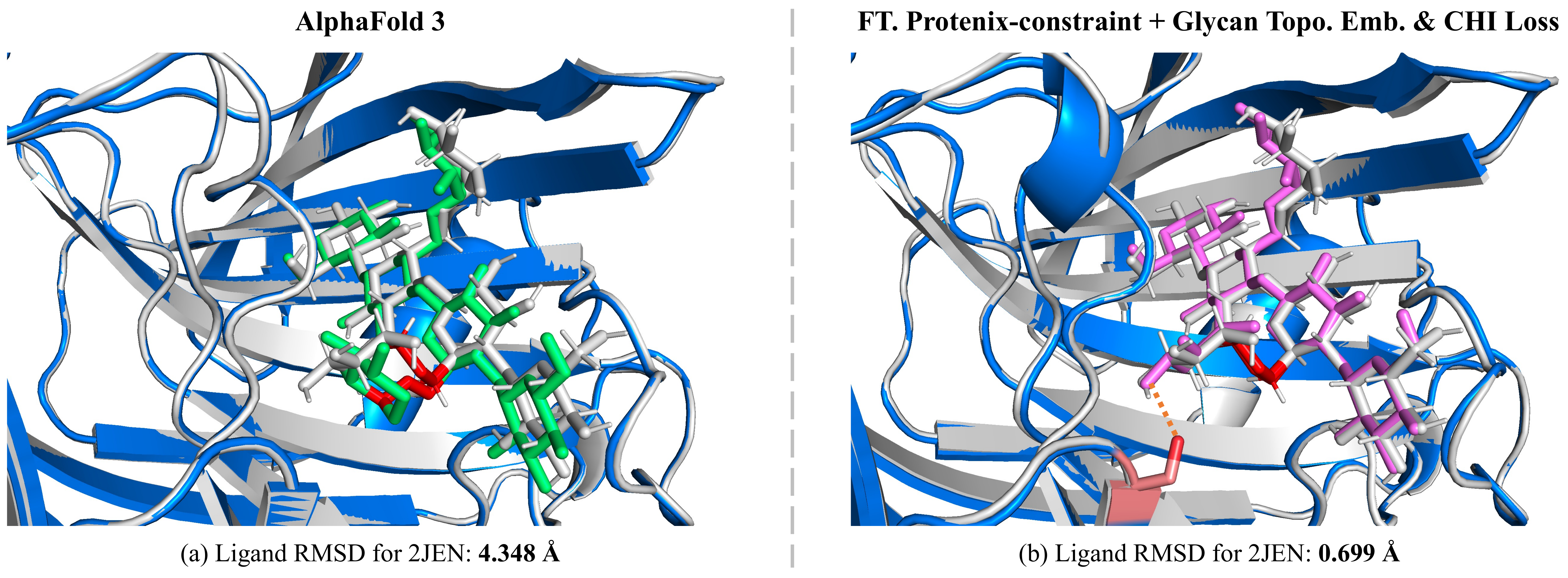}
    \vspace{-1mm}
    \caption{Protein-glycan complex structures predicted by AlphaFold 3 and the finetuned Protenix-constraint with glycan topology embedding and CHI loss. Ground-truth proteins and glycans are shown in \textcolor{gray}{gray}, predicted proteins are shown in {\color[rgb]{0.1445,0.3945,0.8086} blue}, the predicted glycans by AlphaFold 3 show in {\color[rgb]{0.2695,0.8828,0.3672} green}, the predicted glycans by finetuned Protenix-constraint show in {\color[rgb]{0.7695,0.4258,0.9961} purple}, \textcolor{orange}{orange dashed lines} denote hydrogen bonds, and \textcolor{red}{red bonds in glycans} denote the glycosidic bonds leading to prediction conformational differences. \emph{Abbr.}, FT.: Finetuned; Topo.: Topology; Emb.: Embedding.}
    \label{fig:case}
    \vspace{-1.5mm}
\end{figure*}

\textbf{Cases of prediction improvements.} In Figure~\ref{fig:case}, we visualize the predictions of AlphaFold 3 and the Protenix-constraint finetuned with the two proposed techniques for the test sample with PDB ID 2JEN, and more cases are provided in Appendix~\ref{sec:app:case}. For this sample, AlphaFold~3 gets a Ligand RMSD of 4.348~\AA, and the finetuned Protenix-constraint gets a Ligand RMSD of 0.699~\AA. The structural difference between the predictions of two models mainly lies in the glycosidic bond colored in red, where the predicted bond conformation by the finetuned Protenix-constraint is close to the ground truth, while the prediction of AlphaFold~3 is not even close. Due to this structural difference, the finetuned Protenix-constraint succeeds in finding an important hydrogen bond responsible for glycan localization in the binding pocket as reported by \citet{gloster2007characterization} (denoted by an orange dashed line in Figure~\ref{fig:case}(b)), while AlphaFold~3 fails to identify this critical hydrogen bond. This improvement is consistent with the intended role of the CHI energy loss during finetuning, which penalizes energetically unfavorable glycosidic torsion angles and encourages conformationally plausible glycan structures. 

\begin{figure*}[t]
    \centering
    \includegraphics[width=0.93\linewidth]{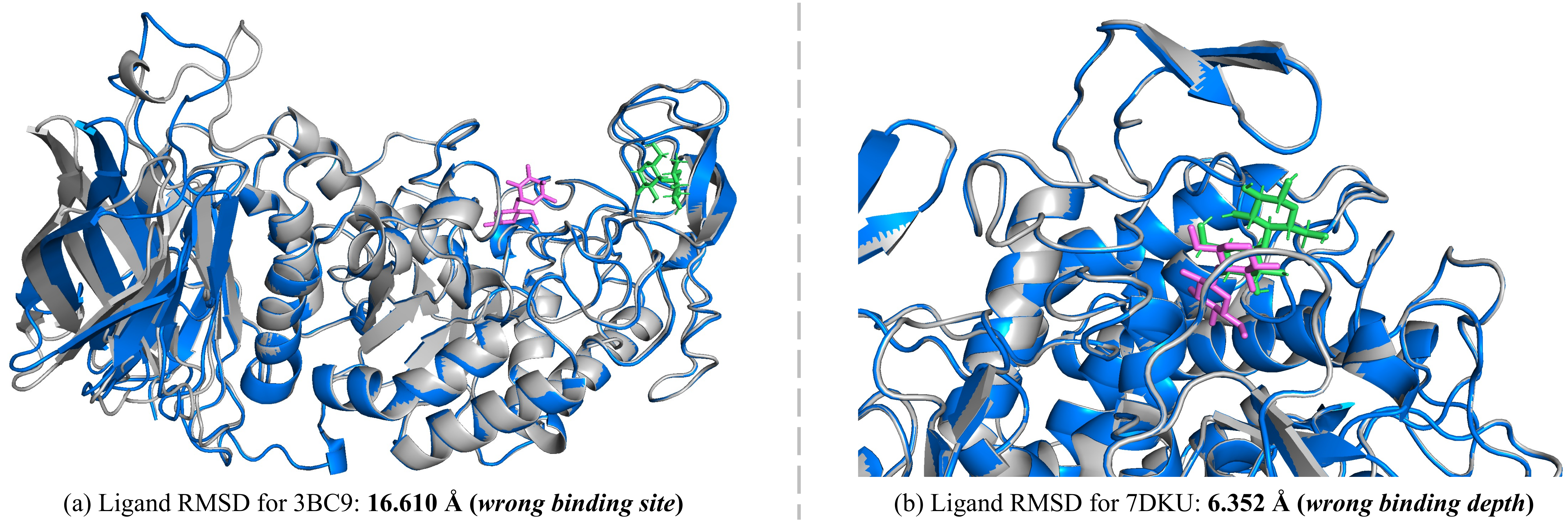}
    \vspace{-1mm}
    \caption{Failure cases of the finetuned Protenix-constraint with glycan topology embedding and CHI loss. Ground-truth proteins show in \textcolor{gray}{gray}, ground-truth glycans show in {\color[rgb]{0.2695,0.8828,0.3672} green}, predicted proteins show in {\color[rgb]{0.1445,0.3945,0.8086} blue}, and predicted glycans show in {\color[rgb]{0.7695,0.4258,0.9961} purple}.}
    \label{fig:fail_case}
    \vspace{-1.5mm}
\end{figure*}

\textbf{Failure cases.} In Figure~\ref{fig:fail_case}, we show two failure cases of the Protenix-constraint finetuned with the two proposed techniques, representing two typical failure modes, and more failure cases are provided in Appendix~\ref{sec:app:case}. For the 3BC9 sample, the model fails to locate the correct glycan-binding site, placing the glycan in a region that does not correspond to the experimentally observed interface. For the 7DKU sample, the model predicts a deeply bound glycan conformation, while the ground-truth glycan should be more shallowly bound, since this glycan, \emph{i.e.}, Cellobiose, serves as a glycosyl donor and should leave sufficient binding space for its acceptor. Together, these cases reveal two complementary challenges: locating the correct binding site and positioning the glycan in a manner consistent with its biological role. They suggest that further improvements may require both more accurate modeling of protein-glycan recognition and the incorporation of functional context. 

\section{Conclusions and Future Work} \label{sec:conclusion}

This work takes the initiative of boosting more accurate protein-glycan docking with a tailored benchmark and a set of customized models. We first curate the high-quality {\benchmark} benchmark to evaluate models' generalizability to novel glycans and proteins. On the benchmark, we evaluate two physics-based docking methods, two machine-learning-based docking models and six molecular structure prediction models. We further propose two finetuning techniques to enhance the ability of general-purpose molecular structure prediction models on protein-glycan docking. According to benchmark results, AlphaFold~3 achieves superior performance over other existing models, and the two proposed finetuning techniques can be well combined to achieve top performance. 

\textbf{Limitations and future work.} The current benchmark comprehensively encompasses existing experimental protein-glycan complex structures, while its data quantity is still not large enough. To further enlarge the dataset scale, we will try to synthesize large-scale protein-glycan complexes through cross-docking in the future. 
Moreover, the current benchmark focuses on the blind docking setting. In future work, we will extend it to protein-glycan docking with partially known protein receptor apo structures and binding sites. 



\newpage


\section*{Acknowledgments} \label{sec:ack}

This work is supported by Fundamental and Interdisciplinary Disciplines Breakthrough Plan of the Ministry of Education of China (JYB2025XDXM113), Beijing Municipal Natural Science Foundation (Grant No. L2603008), National Natural Science Foundation of China (92470121, 62402016), National Key R\&D Program of China (2024YFA1014003), Zhongguancun Academy (C20250204, C20250602),  Beijing Major Science and Technology Project (Z251100008125043, Z251100008425023), and High-performance Computing Platform of Peking University. This work is also supported by the Tencent Project Up Scholarship project. 


\bibliographystyle{plainnat} 
\bibliography{ref.bib}


\newpage
\appendix

\section{More Dataset Statistics} \label{sec:app:statistics}

\begin{figure}[h]
\centering
    \vspace{-2mm}
    \includegraphics[width=1.0\linewidth]{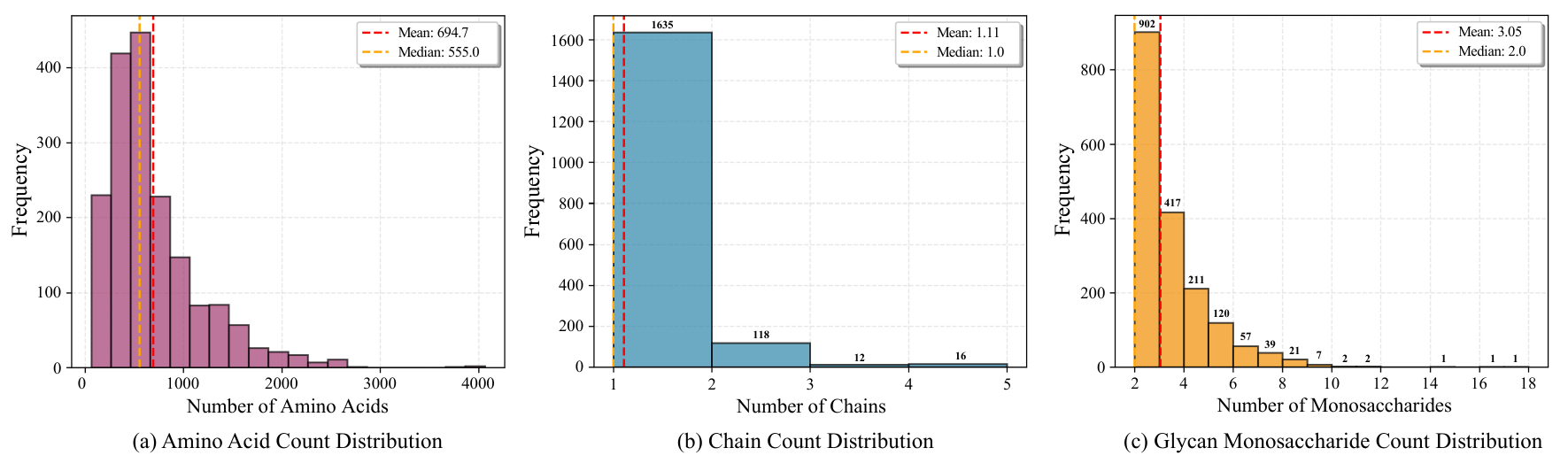}
    \vspace{-6mm}
    \caption{Dataset statistics of protein and glycan sizes. We report the count distribution of amino acid, protein chain and glycan monosaccharide by histogram.}
    \label{fig:statistics}
\end{figure}


\begin{table*}[h]
\vspace{-2.5mm}
\centering
\caption{Dataset statistics of glycan attributes. \emph{Abbr.}, Glc: Glucose; Man: Mannose; GlcNAc: N-Acetylglucosamine; Gal: Galactose; Xyl: Xylose; Ac: Acetylation; Sul: Sulfation; Phos: Phosphorylation; Me: Methylation.} 
\label{tab:glycan_attribute}
\begin{spacing}{1.3}
\vspace{-1mm}
\begin{adjustbox}{max width=1\linewidth}
\begin{tabular}{l|cc|ccccc|cc|cc|ccccc}
\toprule
\multirow{2}{*}{\bf{Attribute}} & \multicolumn{2}{c|}{\bf{Configuration}} & \multicolumn{5}{c|}{\bf{Top-5 Stem Type}} & \multicolumn{2}{c|}{\bf{Ring Size}} & \multicolumn{2}{c|}{\bf{Anomeric Notation}} & \multicolumn{5}{c}{\bf{Top-5 Modification Group}} \\
\cmidrule{2-17}
& \bf{D-config} & \bf{L-config} & \bf{Glc} & \bf{Man} & \bf{GlcNAc} & \bf{Gal} & \bf{Xyl} & \bf{Pyranose} & \bf{Furanose} & \bf{Alpha} & \bf{Beta} & \bf{No modify} & \bf{Ac} & \bf{Sul} & \bf{Phos} & \bf{Me} \\
\midrule
\bf{Percentage} & 98.3\% & 1.7\% & 31.6\% & 17.3\% & 15.7\% & 10.4\% & 7.2\% & 91.6\% & 8.4\% & 47.2\% & 52.8\% & 38.3\% & 24.0\% & 15.6\% & 8.9\% & 6.1\% \\
\bottomrule
\end{tabular}
\end{adjustbox}
\end{spacing}
\end{table*}


\textbf{Statistics of protein and glycan sizes.} Figure \ref{fig:statistics} provides additional statistics of the proposed dataset. As shown in Figure~\ref{fig:statistics}(a), the distribution of amino acid counts per sample is right-skewed, with a mean of 694.7 and a median of 555.0, indicating substantial variability in protein size across samples.

As illustrated in Figure~\ref{fig:statistics}(b), regarding protein complex composition, the dataset is dominated by single-chain proteins. Specifically, 1,635 samples contain a single protein chain, while a small fraction involves multiple chains (118 samples with two chains, 12 samples with three chains, and 16 samples with four chains). 

Figure~\ref{fig:statistics}(c) summarizes the distribution of glycan sizes measured by the number of monosaccharides. Most glycans are relatively small, with the majority consisting of 2 to 4 monosaccharides (902 samples with 2 monosaccharides, 417 samples with 3 monosaccharides, and 211 samples with 4 monosaccharides). The overall mean monosaccharide count is 3.05, while the median is 2.0, indicating a bias toward short glycan chains. Nevertheless, the dataset also includes a long tail of larger glycans, with sizes extending up to 17 monosaccharides, providing coverage of more complex glycan structures. 

\textbf{Statistics of glycan attributes.} In Table~\ref{tab:glycan_attribute}, we report the data distribution on five core attributes of glycans. We observe that: (1) D-configuration dominates the dataset as in nature; (2) Five stem types occupy 82.2\% of all data; (3) Pyranose appears much more frequently than furanose as in nature; (4) Alpha and Beta anomeric notations are balanced; (5) Unmodification and four modification groups occupy 92.9\% of all data. Therefore, the dataset faithfully reflects naturally occurring glycan properties, featuring dominant natural configurations, concentrated mainstream backbone and modification types, and balanced anomeric distribution. 


\section{Setups of Baseline Models} \label{sec:app:baseline}

We set up the 10 baseline models used in this work as below: 
\begin{itemize}
\vspace{-2.3mm}
    \item \textbf{AutoDock Vina} (Apache license): We follow the official configuration of AutoDock Vina v1.2. Under the blind docking setting, following previous works~\citep{corso2022diffdock,morehead2026assessing}, the protein receptor structures are predicted by AlphaFold~3, glycan 3D conformations are generated from SMILES sequences using RDKit~\citep{landrum2013rdkit}, and the binding pockets predicted by DeepPocket~\citep{aggarwal2021deeppocket} are used to define search boxes. 
    \vspace{-0.6mm}
    \item \textbf{Vina-Carb} (Apache license): We employ Vina-Carb 1.0, while keeping all other configurations and the docking pipeline identical to AutoDock Vina. 
    \vspace{-0.6mm}
    \item \textbf{DiffDock} (MIT license): We use the official model checkpoints and its default blind-docking inference settings with AlphaFold~3-predicted receptor structures and RDKit-generated glycan conformations. The generated binding poses are ranked by the confidence model, with the top-ranked pose selected for evaluation.
    \item \textbf{SurfDock} (MIT license): We adopt the official model checkpoints and inference pipeline with AlphaFold~3-predicted receptor structures and RDKit-generated glycan conformations. Protein surface features are computed using the official preprocessing pipeline, and the top-ranked pose scored by SurfScore is selected for evaluation. 
    \item \textbf{RoseTTAFold All-Atom} (BSD license): Following the official guideline, we input the model with the amino acid sequence of a protein and the 3D conformation of a glycan in an sdf file.  
    \vspace{-0.6mm}
    \item \textbf{AlphaFold 3} (CC BY-NC-SA 4.0 license): At the input, a protein is represented by its amino acid sequence, and a glycan is represented as a set of monosaccharides (denoted by CCD codes) with covalent bonds. We take the output sample with highest ranking score as the prediction. 
    \vspace{-0.6mm}
    \item \textbf{Chai-1} (Apache license): We follow the official setup of Chai-1 v0.6.1 with a protein sequence and a glycan SMILES sequence as inputs, outputting the structure with highest ranking score.  
    \vspace{-0.6mm}
    \item \textbf{Boltz-2} (MIT license): We follow Boltz-2 v2.2.1 to set up the model, and its inference pipeline is basically the same as AlphaFold 3. 
    \vspace{-0.6mm}
    \item \textbf{Protenix} (Apache license): We adopt the Protenix v0.5.0 base checkpoint. The input configuration and output scheme of Protenix are identical to AlphaFold 3. 
    \vspace{-0.6mm}
    \item \textbf{Protenix-constraint} (Apache license): We adopt the Protenix v0.5.3 constraint checkpoint. Besides the inputs of Protenix, Protenix-constraint uses the predicted binding pocket from DeepPocket as its pocket constraints. Its output scheme is the same as Protenix. 
\vspace{-2mm}
\end{itemize}


\section{Sensitivity Analysis on CHI Loss Weight} \label{sec:app:sensitivity}

\begin{table*}[h]
\centering
\vspace{-2mm}
\caption{Sensitivity analysis on CHI loss weight $\gamma$. Mean (standard deviation) of each metric is reported. The best result is denoted by \textbf{bold}. \emph{Abbr.}, PB-valid: passing all PoseBusters tests, FT.: Finetuned; Ptx-C.: Protenix-constraint; CL.: CHI Loss.} 
\label{tab:sensitivity}
\begin{spacing}{1.3}
\vspace{-1.5mm}
\begin{adjustbox}{max width=1\linewidth}
\begin{tabular}{l|cccccc|cccccc}
\toprule
\multirow{4}{*}{\bf{\large Model}} & \multicolumn{6}{c|}{\bf{\large Protein-holdout test (148 samples)}} & \multicolumn{6}{c}{\bf{\large Glycan-holdout test (134 samples)}} \\
\cmidrule{2-13}
& \multirow{3}{*}{ \makecell[c]{\bf{\#Ligand} \\ \bf{RMSD $<2$\AA $\uparrow$}}} &
\multirow{3}{*}{ \makecell[c]{\bf{\#Ligand} \\ \bf{RMSD $<2$\AA} \\ \bf{\& PB-valid $\uparrow$}}} & 
\multirow{3}{*}{ \makecell[c]{\bf{Ligand} \\ \bf{RMSD $\downarrow$}}} &
\multirow{3}{*}{ \makecell[c]{\bf{Ring} \\ \bf{sRMSD $\downarrow$}}} &
\multirow{3}{*}{ \makecell[c]{\bf{Complex} \\ \bf{RMSD $\downarrow$}}} &
\multirow{3}{*}{ \makecell[c]{\bf{Contact} \\ \bf{Recovery\% $\uparrow$}}} &
\multirow{3}{*}{ \makecell[c]{\bf{\#Ligand} \\ \bf{RMSD $<2$\AA $\uparrow$}}} &
\multirow{3}{*}{ \makecell[c]{\bf{\#Ligand} \\ \bf{RMSD $<2$\AA} \\ \bf{\& PB-valid $\uparrow$}}} & 
\multirow{3}{*}{ \makecell[c]{\bf{Ligand} \\ \bf{RMSD $\downarrow$}}} &
\multirow{3}{*}{ \makecell[c]{\bf{Ring} \\ \bf{sRMSD $\downarrow$}}} &
\multirow{3}{*}{ \makecell[c]{\bf{Complex} \\ \bf{RMSD $\downarrow$}}} &
\multirow{3}{*}{ \makecell[c]{\bf{Contact} \\ \bf{Recovery\% $\uparrow$}}} \\
&  &  &  &  &  &  &  &  &  &  &  &  \\
&  &  &  &  &  &  &  &  &  &  &  &  \\
\midrule

FT. Ptx-C. + CL. ($\gamma = 0.001$) & 36.2 (0.7) & 24.2 (0.7) & 15.54 (0.38) & 0.96 (0.05) & 8.17 (0.30) & 39.5 (1.0) & 35.8 (0.4) & 24.0 (0.6) & 12.98 (0.47) & 1.22 (0.06) & 3.94 (0.25) & 39.1 (0.9) \\
FT. Ptx-C. + CL. ($\gamma = 0.01$)  & \textbf{38.6} (0.5) & \textbf{25.8} (0.4) & \textbf{14.61} (0.53) & \textbf{0.86} (0.05) & \textbf{7.37} (0.26) & \textbf{42.3} (0.7) & \textbf{37.0} (0.6) & \textbf{25.6} (0.8) & \textbf{11.82} (0.33) & \textbf{1.10} (0.05) & \textbf{3.26} (0.16) & \textbf{42.0} (0.8) \\
FT. Ptx-C. + CL. ($\gamma = 0.1$)   & 35.4 (0.5) & 23.6 (0.5) & 16.07 (0.41) & 0.98 (0.06) & 8.53 (0.37) & 39.1 (0.8) & 34.8 (0.4) & 23.2 (0.7) & 13.25 (0.43) & 1.28 (0.10) & 4.27 (0.23) & 38.3 (1.1) \\
FT. Ptx-C. + CL. ($\gamma = 0.5$)   & 31.0 (0.6) & 20.6 (0.8) & 16.79 (0.45) & 1.03 (0.09) & 9.44 (0.39) & 37.6 (0.7) & 31.6 (0.5) & 20.8 (0.4) & 13.75 (0.38) & 1.32 (0.11) & 4.43 (0.26) & 37.5 (0.9) \\
\bottomrule
\end{tabular}
\end{adjustbox}
\end{spacing}
\vspace{2.5mm}
\end{table*}


In Table~\ref{tab:sensitivity}, we report the performance of the finetuned Protenix-constraint with CHI loss under the CHI loss weight 0.001, 0.01, 0.1 and 0.5. It can be observed that, on both test sets, the model with the CHI loss weight 0.01 performs best in terms of RMSD metrics, contact recovery rate, and the number of geometrically valid and physically plausible docking samples. This result is attributed to the fact that the CHI loss weight 0.01 makes the CHI loss and the original structure prediction loss at a similar scale, which well balances two tasks, glycan conformation refinement and complex structure prediction. 


\section{More Case Studies} \label{sec:app:case}

\begin{figure*}[h]
    \centering
    \includegraphics[width=0.90\linewidth]{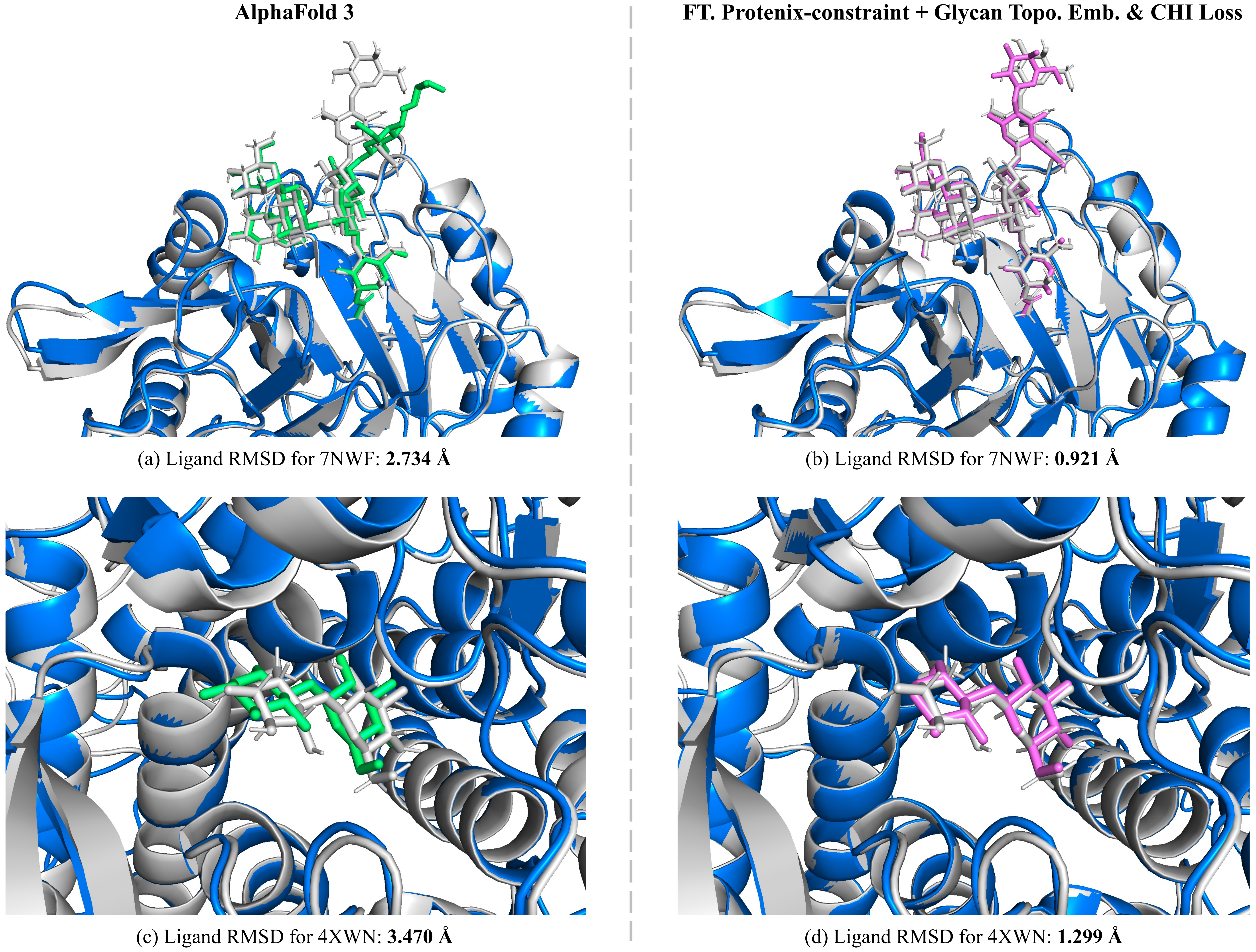}
    \caption{Visualization of predicted protein-glycan complex structures by AlphaFold 3 and the finetuned Protenix-constraint with glycan topology embedding and CHI loss. Ground-truth proteins and glycans are shown in \textcolor{gray}{gray}, predicted proteins are shown in {\color[rgb]{0.1445,0.3945,0.8086} blue}, the predicted glycans by AlphaFold 3 show in {\color[rgb]{0.2695,0.8828,0.3672} green}, and the predicted glycans by finetuned Protenix-constraint show in {\color[rgb]{0.7695,0.4258,0.9961} purple}. \emph{Abbr.}, FT.: Finetuned; Topo.: Topology; Emb.: Embedding.} 
    \label{fig:supp_case}
    \vspace{-1mm}
\end{figure*}


\begin{figure*}[h]
    \centering
    \includegraphics[width=1.0\linewidth]{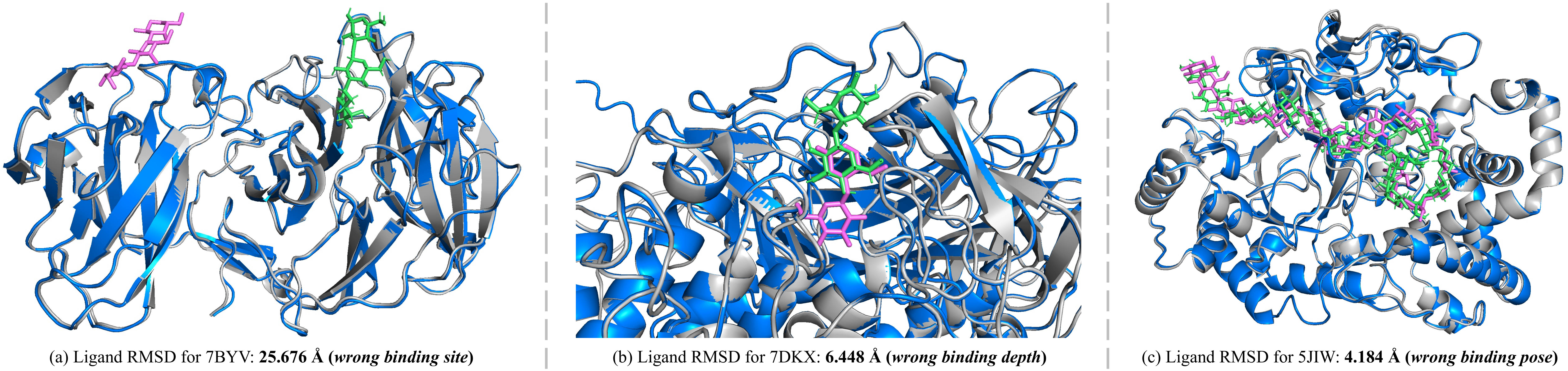}
    \caption{Failure cases of the finetuned Protenix-constraint with glycan topology embedding and CHI loss. Ground-truth proteins show in \textcolor{gray}{gray}, ground-truth glycans show in {\color[rgb]{0.2695,0.8828,0.3672} green}, predicted proteins show in {\color[rgb]{0.1445,0.3945,0.8086} blue}, and predicted glycans show in {\color[rgb]{0.7695,0.4258,0.9961} purple}.}
    \label{fig:supp_fail_case}
\end{figure*}


\textbf{More cases of prediction improvements.} In Figure~\ref{fig:supp_case}, we visualize two more predictions of AlphaFold~3 and the Protenix-constraint finetuned with the two proposed techniques. On the test samples with PDB ID 7NWF and 4XWN, the finetuned Protenix-constraint all achieves a lower Ligand RMSD than AlphaFold~3. By investigating the predicted structures, it can be observed that the glycan binding poses predicted by the finetuned Protenix-constraint is obviously closer to the ground truth than AlphaFold~3 predictions. These results again demonstrate the effectiveness of the proposed finetuning techniques in refining the glycan conformation in a protein binding pocket. 

\textbf{More failure cases.} In Figure~\ref{fig:supp_fail_case}, we present three more failure cases of the Protenix-constraint finetuned with the two proposed techniques. In these three cases, the model respectively fails to identify the glycan binding site, docks the glycan overly deep inside the binding pocket, and predicts a wrong glycan binding pose. These represent three typical failure modes of the model, which we will prioritize addressing in our future work.


\section{Broader Impacts} \label{sec:app:impact}

This work aims to build a tailored benchmark and develop a set of customized models for protein-glycan docking. The proposed models have the potential to help real-world applications related to protein-glycan interactions, such as synthetic biology~\citep{kightlinger2020synthetic} and vaccine design~\citep{kaplonek2018improving}. 

However, we should not ignore the potential risks brought by the proposed models, \emph{e.g.}, developing a novel microorganism disrupting local ecosystems. Therefore, we will strive to ensure the responsible deployment of the proposed models in real-world applications. 




\end{document}